\documentclass[runningheads]{llncs}
\usepackage[T1]{fontenc}
\usepackage{graphicx}
\usepackage{cite}
\usepackage{amsmath,amssymb,amsfonts}
\usepackage{algorithmic}
\usepackage{textcomp}
\usepackage{xcolor}
\usepackage{diagbox}
\begin{document}
\title{Generalized Rapid Action Value Estimation in Memory-Constrained Environments}
\titlerunning{GRAVE in Memory-Constrained Environments}
% If the paper title is too long for the running head, you can set
% an abbreviated paper title here
%

\author{Aloïs Rautureau\inst{1}\orcidID{0009-0007-8507-2167} \and
Tristan Cazenave\inst{2}\orcidID{000-0003-4669-9374} \and Éric Piette\inst{1}\orcidID{0000-0001-8355-636X}}
\authorrunning{A. Rautureau et al.}
% First names are abbreviated in the running head.
% If there are more than two authors, 'et al.' is used.
%
\institute{ICTEAM, UCLouvain, Louvain-la-Neuve, Belgium\and
LAMSADE, Université Paris-Dauphine, Paris, France}
\maketitle              % typeset the header of the contribution
\begin{abstract}
Generalized Rapid Action Value Estimation (GRAVE) has been shown to be a strong variant within the Monte-Carlo Tree Search (MCTS) family of algorithms for General Game Playing (GGP). However, its reliance on storing additional win/visit statistics at each node makes its use impractical in memory-constrained environments, thereby limiting its applicability in practice. In this paper, we introduce the GRAVE$^2$, GRAVER and GRAVER$^2$ algorithms, which extend GRAVE through two-level search, node recycling, and a combination of both techniques, respectively. We show that these enhancements enable a drastic reduction in the number of stored nodes while matching the playing strength of GRAVE.

\end{abstract}

\section{Introduction}
Monte-Carlo Tree Search (MCTS) \cite{coulomEfficientSelectivityBackup2006} is a family of asymmetric partial tree search algorithms that have proven successful in a wide range of decision-making tasks, even in domains where domain-specific knowledge is scarce or difficult to encode symbolically. These methods, however, are typically developed under the implicit assumption that sufficient memory is available to store an ever-growing asymmetric search tree. While this assumption is largely valid on most modern desktop and server-class hardware, it significantly limits the applicability of MCTS-based agents on memory-constrained platforms such as microcontrollers or smartphones.

This limitation is particularly pronounced for Generalized Rapid Action Value Estimation (GRAVE) \cite{cazenaveGeneralizedRapidAction}, which improves the playing strength of agents based on MCTS in several games by augmenting each node with additional win/visit statistics for the All-Moves As First (AMAF) selection policy \cite{helmboldAllMovesAsFirstHeuristicsMonteCarlo2009}. These additional statistics introduce a constant-factor increase in node memory usage, making GRAVE particularly sensitive to memory constraints. As a result, enabling GRAVE-level performance while drastically reducing the number of stored nodes becomes a key challenge for deploying strong artificial agents in environments where memory is a scarce resource.

Beyond practical deployment concerns, this challenge also aligns naturally with questions in cognitive modeling \cite{rautureau2025cogniplayworkinprogresshumanlikemodel}. Best-first tree search algorithms have been proposed as models of human forward planning and short-term memory \cite{degrootThoughtChoiceChess1946}, yet it is unrealistic to assume that human decision makers can retain more than a few dozen decision states simultaneously \cite{atkinsonHumanMemoryProposed1968} \cite{millerMagicalNumberSeven1956}. From this perspective, limiting the number of stored nodes is not merely an engineering constraint but a desirable modeling property. Developing GRAVE-like algorithms that preserve strong best-first behavior and playing strength while operating under a strict bound on the number of nodes, therefore, contributes both to the design of efficient agents and to the algorithmic modeling of high-level features of human decision making in games. In this setting, nodes are treated as atomic units of information rather than as byte-encoded data structures, shifting the focus from raw memory usage to principled control over informational complexity.

To address this challenge, we introduce GRAVE$^2$ and GRAVER (Generalized Rapid Action Value Estimation with node Recycling), two variants of GRAVE designed for low memory budgets while matching the playing strength of the original algorithm. GRAVE$^2$ extends GRAVE using a two-level search tree approach \cite{breukerMemorySearchGames1998}, while GRAVER incorporates a node recycling scheme \cite{powleyMemoryBoundedMonte2017}. We further propose GRAVER$^2$, which combines the two-level search of GRAVE$^2$ with the node recycling mechanism of GRAVER. To our knowledge, this is the first implementation of an MCTS algorithm that integrates both node recycling and two-level search within a single framework.

We evaluate these algorithms against GRAVE and corresponding UCT variants, comparing their relative playing strength under strict memory constraints. In particular, we measure the number of stored nodes required to reach equal playing strength, the number of playouts performed, and the empirically measured memory footprint of our implementation. The results provide insight into the trade-offs between memory usage and performance, and open new perspectives for the development of memory-bounded MCTS methods.

\section{Related work} \label{sect:related_work}

MCTS algorithms typically differ in their selection and playout policies. The standard selection policy, UCT, balances exploration of the search tree with exploitation of nodes with high estimated value using the Upper Confidence Bound (UCB) formula:
\begin{equation}
\arg\max_n (\frac{R(n)}{V(n)} + C \sqrt{\frac{\log V(N)}{V(n)}}
\end{equation}
where $n$ is a node, $N$ its parent, $V$ and $R$ denote the number of visits and the mean sampled reward of a node, respectively, and $C$ is the exploration parameter. Low values of $C$ favor exploitation, while high values encourage exploration. The value $C = \frac{\sqrt{2}}{2}$ is known to provide good results when sampled rewards lie in the range $[0, 1]$ \cite{kocsisBanditBasedMonteCarlo2006}, and is used throughout this paper.

GRAVE \cite{cazenaveGeneralizedRapidAction} is a generalization of the Rapid Action Value Estimation (RAVE) \cite{gellyMonteCarloTreeSearch2011} algorithm. It relies on the All-Moves As First (AMAF) heuristic \cite{bouzy2003MonteCarloDevelopments,helmboldAllMovesAsFirstHeuristicsMonteCarlo2009} as part of its selection policy, and linearly interpolates between the exploitation of the mean sampled reward of a node $n$ and the AMAF value of a move $m$ stored in a reference node $ref$ higher up the tree. The interpolation is controlled by a parameter $\beta_{n, m}$ defined as:

\begin{equation}
\beta_{n, m} = \frac{AMAF_{ref, m}}{AMAF_{ref, m} + V(n) + (bias \times AMAF_{ref, m} \times V(n))}
\end{equation}

We then select the most promising child node using the formula:

\begin{equation}
    \arg\max_n ((1 - \beta_n) \times \frac{R(n)}{V(n)} + \beta_n \times AMAF_{ref, m})
\end{equation}

The bias parameter controls the extent to which AMAF heuristics are favored over direct exploitation of node values, while a reference threshold specifies the minimum number of visits required for a node’s AMAF statistics to be reused deeper in the search.

GRAVE has been shown to outperform traditional UCT and RAVE in a number of games with permutable moves, benefiting from the additional information provided by AMAF heuristics. In domains where strong heuristics are not available, such as General Game Playing (GGP), it often constitutes a stronger alternative to other MCTS algorithms \cite{pietteNouvelleApprocheAu2016,sironiComparisonRapidAction2016,Koriche_2017_Symmetry,Koriche_2017_Woodstock}.

Like most MCTS variants, GRAVE requires storing the explored search tree in memory to maintain value estimates for each node. This paper focuses on two complementary approaches for reducing the size of the stored tree while minimizing the impact on playing strength.

Two-level search algorithms offer one such approach, effectively squaring the number of playouts performed while increasing the number of stored nodes by only a constant factor of two. Similar techniques have been successfully applied to Proof-Number (PN) Search, where they are known as PN$^2$\cite{allisSearchingSolutionsGames1994,breukerMemorySearchGames1998}.

These algorithms operate by performing a standard search from the root node (\textit{top-level search}), but replace the usual leaf expansion step with a second search initiated from the newly expanded leaf node (\textit{second-level search}). The second-level search tree is then discarded, and the root of this secondary search is added as a leaf to the top-level tree, with the collected values propagated upward, typically in a batch. This approach allows the algorithm to gather more accurate estimates before backpropagation, while enabling the reuse of a large fraction of the available node budget.

While two-level search algorithms typically split the node budget evenly between the two levels, they can be generalized by introducing a parameter $\lambda \in [0, 1]$ that determines the fraction of the node budget allocated to the second-level search tree. Let $N_{sec} = N * \lambda$ and $N_{top} = N - N_{sec}$ denote the number of nodes assigned to the second- and top-level trees, respectively. A two-level search with total node budget $N$ and parameter $\lambda$ will therefore perform $(\lambda - \lambda^2) * N^2$ iterations in practice, compared to only $N$ iterations for a single-level search using the same node budget.

Orthogonal to two-level approaches, node recycling was first introduced for Information Set MCTS (ISMCTS) \cite{cowlingInformationSetMonte2012, powleyMemoryBoundedMonte2017}. The core idea is to reuse the memory of nodes that have been least recently accessed in order to allow continued expansion under a fixed memory budget. Such nodes are considered less relevant to the current search, as MCTS algorithms inherently revisit promising nodes more frequently.

In this approach, a fixed pool of $N$ nodes is allocated, initially containing empty entries. While empty entries remain, newly expanded nodes are added to the pool in the same manner as in standard MCTS. Once the pool is full, the least recently used node is recycled and replaced by the newly expanded node. Throughout this paper, we note $P$ the number of playouts performed.

To efficiently identify the least recently accessed node, a Least Recently Used (LRU) cache is maintained during the search. When a node is visited during the selection phase, it is temporarily removed from the LRU and reinserted during backpropagation, in the reverse order of removal. This maintains the crucial invariant that only leaf nodes can appear at the front of the LRU, ensuring that internal nodes are never recycled, which would otherwise render their children unreachable.

\section{Methods} \label{sect:methods}
\begin{figure}
    \centering
    \centerline{\includegraphics[width=0.37\linewidth]{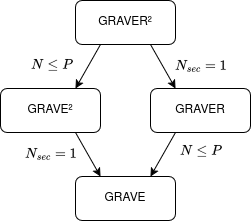}}
    \caption{Relationships between GRAVE, GRAVE$^2$, GRAVER and GRAVER$^2$. $N$ indicates the total number of nodes stored, $P$ the total number of playouts performed, and $N_{sec}$ the nodes stored in the second-level tree.}
    \label{fig:grave_lattice}
\end{figure}

We introduce two variants of the GRAVE algorithm focusing on memory-constrained environments, respectively incorporating two-level search (GRAVE$^2$) and node recycling (GRAVER). These variants are finally combined into GRAVER$^2$, which can utilize node recycling in both the top and second-level trees to further increase the efficient usage of the fixed memory budget.

These variants can be seen as generalizations of one another (Figure \ref{fig:grave_lattice}), providing more parameters to fine tune the playing strength and memory usage of the original GRAVE algorithm.

\subsection{GRAVE$^2$}

GRAVE$^2$ is a two-level adaptation of the GRAVE algorithm that performs a second-level search when expanding leaf nodes.

\begin{figure}
    \centering
    \centerline{\includegraphics[width=0.37\linewidth]{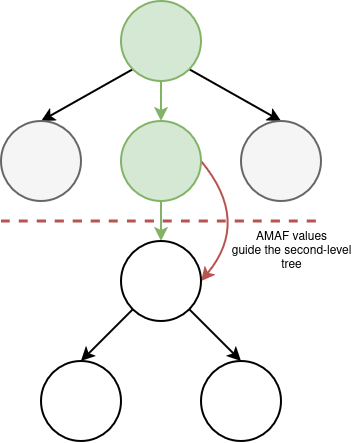}}
    \caption{
    Forward node sharing in GRAVE$^2$. The selection path in the top-level tree is fixed while the second-level search is running, and the latter may use AMAF values aggregated in the top-level tree to guide its exploration as long as the second-level root has fewer visits than the parameterized reference threshold. Values obtained from playouts in the second-level tree are backpropagated to the top-level tree after each iteration, rather than in a single batch once the second-level search terminates. If a child of the currently referenced node exceeds the reference threshold, it becomes the new referenced node.
    }
    \label{fig:forward_sharing}
\end{figure}

Two-level search algorithms typically isolate the top-level and second-level search trees, propagating only the values of the second-level root back to the top-level tree. In the case of GRAVE, however, AMAF values stored in the top-level tree can be reused within the second-level search. We refer to this mechanism as forward sharing (see Figure \ref{fig:forward_sharing}). Forward sharing enables the second-level search to exploit information gathered from previous second-level searches to guide exploration earlier, even though the corresponding trees have been discarded. This property makes GRAVE$^2$ particularly well suited to preserving playing strength while further reducing the algorithm’s memory footprint.

Performing a second-level search before backing up AMAF values produces more reliable estimates, thereby reducing the amount of noise propagated up the tree when expanding a top-level node. By adjusting the $\lambda$ parameter, we can control the trade-off between permanently stored information, holding crucial AMAF statistics for the second-level search, and the quality of information obtained for newly expanded nodes.

In time-bounded scenarios, which represent the most common use case for game-playing agents, two-level search introduces a drawback by limiting the anytime behavior of MCTS algorithms. The search can now only terminate every $\lambda \times N$ playouts rather than after each individual playout. To mitigate this issue and avoid spending computation on unpromising nodes, an early termination mechanism can be employed. A related idea was applied in the PN² algorithm, where the second-level search is halted when the proof number of the second-level root exceeds twice its disproof number \cite{winandsPNPN2PN2001}.

\subsection{GRAVER}
The data structure introduced for node recycling—based on a left-child right-sibling representation of the tree combined with an intrusive first-in first-out LRU cache—can be directly applied to the GRAVE algorithm. We refer to the resulting method as GRAVER (GRAVE with node Recycling) throughout the remainder of this paper.

The main difference with UCT lies in the fact that leaf nodes in GRAVE may contain substantial information in the form of AMAF statistics accumulated from previous playouts, which could prove useful later in the search. Recycling such nodes may therefore have a negative impact on the playing strength of GRAVER. On the other hand, GRAVE’s selection policy can assign meaningful values to unexpanded or recycled nodes through AMAF heuristics. While UCT conventionally assigns a value $UCB(n) = \infty$ to unexpanded nodes, ensuring they are always selected for expansion, GRAVE instead relies solely on stored AMAF values for such nodes, since a visit count of zero forces $\beta_m = 1$. This behavior helps prevent pathological cycles in which the algorithm repeatedly expands and recycles the same nodes.

\subsection{GRAVER$^2$}
Node recycling and two-level search are orthogonal methods, enabling the use of both of them simultaneously. We dub the resulting algorithm GRAVER$^2$. Recycling nodes within the top-level tree further improves memory efficiency at the cost of potentially discarding additional information when leaf nodes are recycled. In the second-level search, additional playouts can improve the quality of value estimates while maintaining a fixed node budget. Empirically, node recycling schemes exhibit a plateau in playing strength as the ratio between performed playouts and stored nodes increases. This allows maximizing the playing strength of the algorithm while bounding both memory usage and the number of playouts in the second-level search.

We use the notation $P_{top}$ (resp. $P_{sec}$) to denote the number of playouts performed in the top-level (resp. second-level) tree. The total number of playouts is now decorrelated from the number of nodes stored and the $\lambda$ parameter, with $P = P_{top} \times P_{sec}$.

\section{Experimental results} \label{sect:results}
We evaluate the relative playing strength of the proposed algorithms on the game of Go using a $9 \times 9$ board. The GRAVE algorithm was first tuned against a UCT baseline with an exploration constant of $0.7$ to determine suitable values for the bias and reference threshold parameters for Go $9 \times 9$. These parameters were fixed to $10^{-2}$ and $25$, respectively, for all subsequent experiments.

All playouts are guided by the Move Average Sampling Technique (MAST) \cite{finnssonLearningSimulationControl2010} using an $\epsilon$-greedy sampling strategy with $\epsilon = 0.4$. MAST statistics are decayed by a factor of $0.2$ between successive turns within the same game.

All algorithms are evaluated against a baseline GRAVE implementation with $P = N = 10{,}000$. Each data point is computed from $500$ games. Win rates are reported together with confidence intervals calculated using the Agresti--Coull method \cite{agresti1998}.

\subsection{Two-level search}
\begin{figure}
    \centerline{\includegraphics[width=0.5\linewidth]{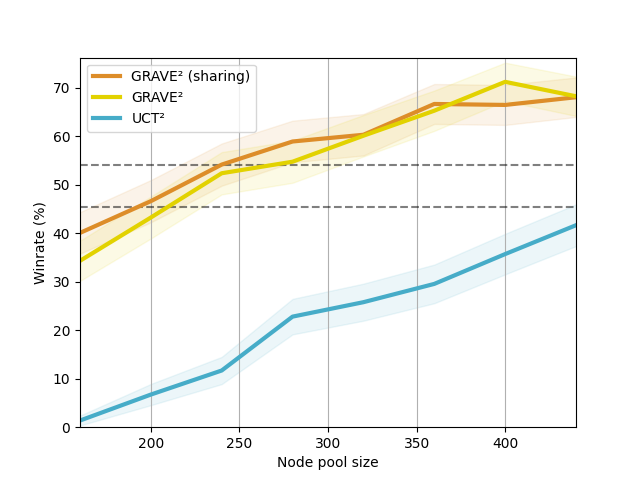}}
    \caption{
    Winrates of GRAVE$^2$ with and without forward sharing and UCT² against GRAVE with $P = N = 10,000$. The dotted lines indicate the 95\% confidence interval of the winrate of GRAVE against itself, representing the region in which compared algorithms can be considered to have playing strength equal to GRAVE.
    }
    \label{fig:two_level_scaling}
\end{figure}

\begin{table}
    \begin{center}
    \begin{tabular}{|c|c|c|c|c|c|}
    \hline
    \diagbox[width=7em]{Nodes}{$\lambda$} & $0.2$ & $0.4$ & $0.5$ & $0.6$ & $0.8$ \\ \hline
    $160$ & $31.7\%$ & $33.7\%$ & $\mathbf{40.0\%}$   & $31.4\%$ & $26\%$ \\ \hline
    $200$ & $41.1\%$ & $\mathbf{48.4\%}$ & $46.6\%$ & $47.2\%$ & $33.8\%$ \\ \hline
    $240$ & $51.8\%$ & $\mathbf{55.6\%}$ & $54.2\%$ & $52.2\%$ & $39.2\%$ \\ \hline
    $280$ & $51.2\%$ & $57.7\%$ & $\mathbf{58.9\%}$ & $52.4\%$ & $43\%$ \\ \hline
    $320$ & $59.3\%$ & $\mathbf{62.7\%}$ & $60.3\%$ & $61.5\%$ & $48.6\%$ \\ \hline
    $360$ & $63.5\%$ & $64.5\%$ & $\mathbf{66.7\%}$ & $63.7\%$ & $53.2\%$ \\ \hline
    $400$ & $62.5\%$ & $\mathbf{69.4\%}$ & $66.5\%$ & $63.9\%$ & $58.4\%$ \\ \hline
    $440$ & $67.5\%$ & $65.3\%$ & $\mathbf{68.1\%}$ & $66.7\%$ & $59.6\%$ \\ \hline
    \end{tabular}
    \label{table:lambda_parameter}
    \end{center}
    \caption{Winrate of GRAVE$^2$ with forward sharing for varying node pool sizes and $\lambda$ values against GRAVE running for $P = 10,000$. Confidence intervals are omitted for clarity, and lie in the range $[4.0\%, 4.4\%]$.}
\end{table}

We first evaluate the performance of GRAVE$^2$ relative to GRAVE. The relative win rates for $\lambda = 0.5$ are reported as a function of the number of nodes stored by the two-level search algorithms in Figure~\ref{fig:two_level_scaling}.

GRAVE$^2$ without forward sharing achieves playing strength comparable to GRAVE starting from a total node pool size of $240$ ($N_{top} = N_{sec} = 120$, corresponding to $P = 14{,}400$). Forward sharing does not yield a statistically significant improvement in overall playing strength. However, its effect is sufficient that we cannot reject the hypothesis of equal playing strength with only $200$ nodes when forward sharing is enabled ($N_{top} = N_{sec} = 100$, corresponding to $P = 10{,}000$). In the remainder of this paper, any reference to GRAVE$^2$ assumes that forward sharing is enabled.

For comparison, UCT$^2$ requires more than $440$ nodes to reach a similar level of playing strength, performing more than four times as many playouts ($P \geq 48{,}400$). It is important to note, however, that storing AMAF statistics in each node results in a substantially larger memory footprint for GRAVE-based methods. In Go $9 \times 9$, each GRAVE node can store up to $82$ AMAF entries (one per intersection plus the pass move), corresponding in our implementation to an additional $82 \times 8 = 656$ bytes per node compared to UCT.

Finally, we assess the impact of the $\lambda$ parameter, which controls the fraction of the node pool allocated to the second-level search, by varying $\lambda$ over the range $[0.2, 0.8]$ for $160 \leq N \leq 440$. The results reported in Table~\ref{table:lambda_parameter} confirm that $\lambda = 0.5$ yields the best overall performance. Reducing $N_{top}$ ($\lambda > 0.5$) generally degrades performance, while reducing $N_{sec}$ ($\lambda < 0.5$) does not produce statistically significant improvements and increases the number of playouts performed.

\subsection{Node recycling}
\begin{figure}
    \centering
    \centerline{\includegraphics[width=0.5\linewidth]{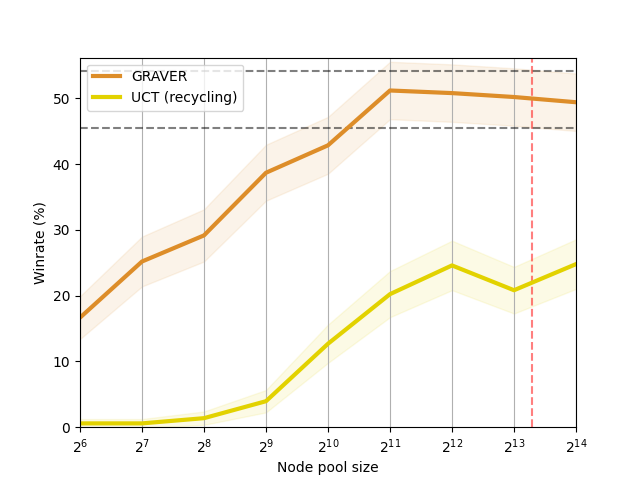}}
    \caption{
    Winrates of GRAVER and UCT with node recycling against GRAVE with $P = 10,000$. The node pool size used is presented on a logarithmic scale. The red dashed line indicates the threshold $N = 10,000$, beyond which all expanded nodes can be stored and node recycling no longer takes effect.
    }
    \label{fig:node_recycling}
\end{figure}

Node recycling is less effective at reducing the memory footprint of the search tree while maintaining equivalent playing strength. GRAVER matches the performance of GRAVE only from approximately $N \approx 1{,}536$, with both algorithms running for $10{,}000$ playouts. 

However, node recycling schemes preserve the anytime property of the MCTS algorithm, which may be crucial in certain scenarios. Detailed comparative win rate results are shown in Figure~\ref{fig:node_recycling}.

\subsection{Node recycling in two-level search}
We evaluate the performance of combining node recycling with two-level search through two separate experiments. In both cases, we set $\lambda = 0.5$ and vary the number of additional playouts as a ratio of the node pool size. A ratio of $1.0$ corresponds to the setting without node recycling.

\begin{figure}
    \centering
    \includegraphics[width=0.48\linewidth]{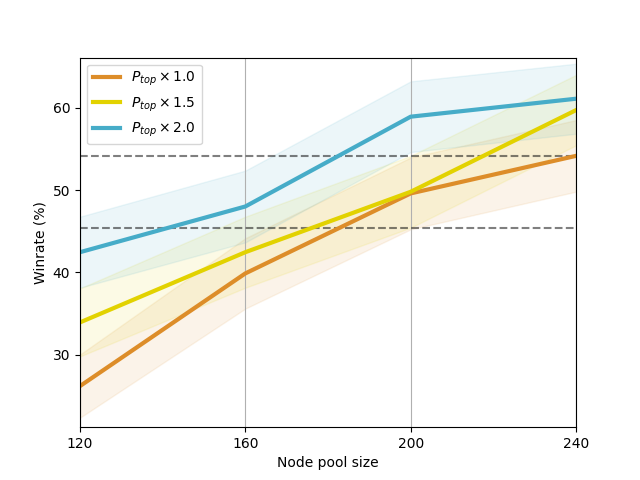}
    \includegraphics[width=0.48\linewidth]{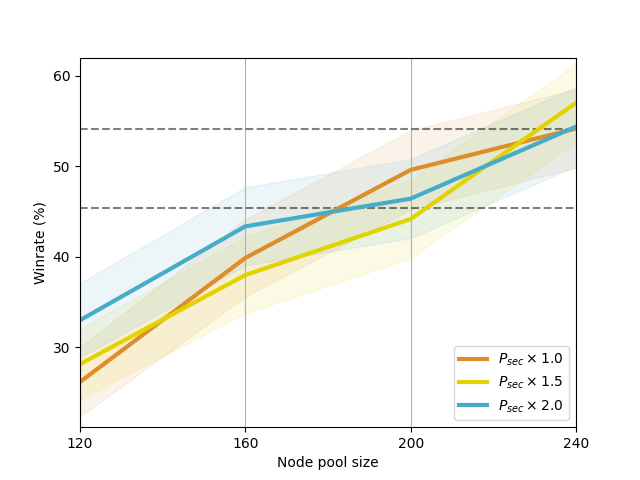}
    \caption{Winrate of GRAVER$^2$ against GRAVE ($P = 10,000$), varying the ratio of playouts to stored nodes in the top-level tree (\textbf{Left}) and second-level tree (\textbf{Right}).}
    \label{fig:graver_sq_top_level}
\end{figure}

We first compare GRAVE and GRAVER$^2$ when node recycling is applied only to the top-level tree (right panel of Figure~\ref{fig:graver_sq_top_level}). Our first observation is that combining two-level search with node recycling improves playing strength under memory constraints. In particular, we are able to reduce the node pool size to $160$ nodes while maintaining performance comparable to baseline GRAVE.

This threshold is noteworthy because it prevents the tree from expanding all legal moves at the root while still reaching depth $\geq 1$. This indicates that the algorithm effectively prunes less promising branches in order to concentrate its limited memory budget on more relevant parts of the search space.

We then examine the effect of applying node recycling within the second-level tree (right panel of Figure~\ref{fig:graver_sq_top_level}). The impact on playing strength is not statistically significant, except at very small node pool sizes ($N \leq 160$).

\begin{figure}
    \centering
    \includegraphics[width=0.48\linewidth]{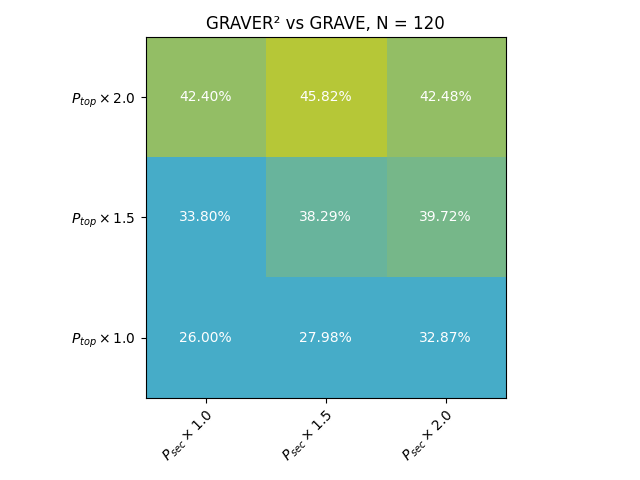} \includegraphics[width=0.48\linewidth]{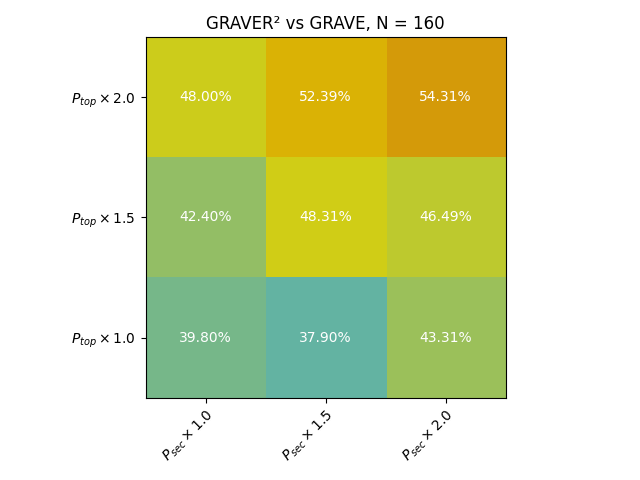} 
    
    \includegraphics[width=0.48\linewidth]{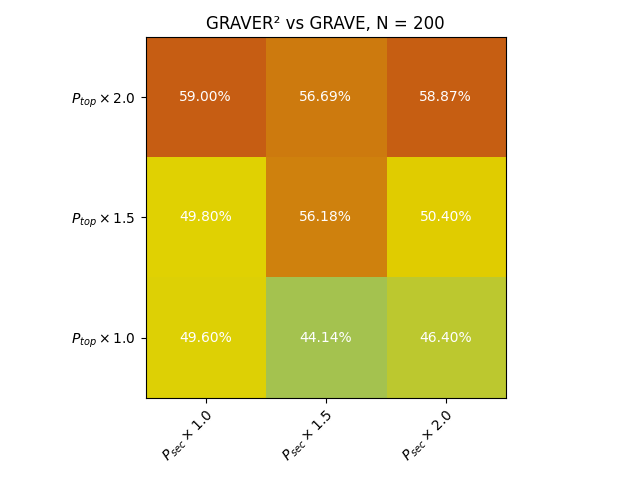} \includegraphics[width=0.48\linewidth]{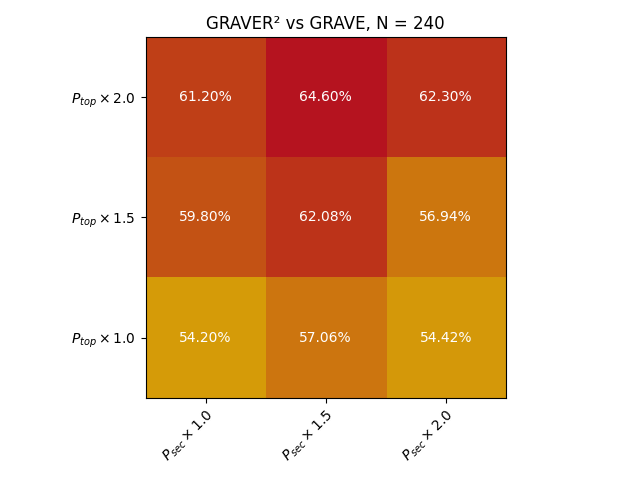}
    \caption{Winrate of GRAVER$^2$ against GRAVE ($P = 10,000$), varying the ratio of playouts to stored nodes in both the top and second-level trees. The values are given with a confidence interval of $\pm 4.2$.}
    \label{fig:graver_sq_both}
\end{figure}

Finally, we evaluate the performance of GRAVER$^2$ by varying the ratio of playouts to stored nodes in both the top- and the second-level trees (Figure \ref{fig:graver_sq_both}). Increasing $P_{sec}$ appears to be less effective than increasing $P_{top}$, and may even degrade playing strength for larger values of $N$. 

One possible explanation is that results backpropagated from simulations guide by MAST exhibit high variance, which can negatively affect estimate quality as the number of playouts increases. In this setting, the top-level tree may not accumulate sufficient information to compensate for the loss of finer-grained statistics in the second-level tree, leading to reduced overall performance.

\section{Conclusion} \label{sect:conclusion}

We presented two orthogonal approaches to reduce the memory usage of GRAVE while maintaining playing strength in Go $9 \times 9$. Two-level search (GRAVE$^2$) achieves the largest reduction, reaching equivalent performance with as little as $2\%$ of the original node count. Node recycling (GRAVER) also reduces the node pool size, though more moderately (approximately $15.36\%$), while preserving the anytime property of MCTS.

Combining both approaches (GRAVER$^2$) yields an even greater reduction in stored nodes, achieving the performance of GRAVE with $N = 160$, $P_{top} = 160$ and $P_{sec} = 80$, corresponding to $12,800$ total playouts. This configuration reduces memory usage to approximately $1.6\%$ while maintaining equivalent playing strength. Applying node recycling within the second-level tree does not produce a statistically significant improvement beyond these results.

Within the scope of our experiments, we also observe that a sharing factor of $\lambda = 0.5$ in two-level search provides the best overall performance. Deviating from this value either reduces playing strength or increases the number of playouts without yielding meaningful strength gains.

\subsection{Future work}
Although this work substantially reduces the memory usage of GRAVE and introduces several tunable parameters, further reductions may be possible through the integration of stronger heuristics and heuristic-based pruning strategies \cite{moriartyEvolvingNeuralNetworks1994}. Such approaches could shift the balance toward search-light or partially searchless methods \cite{mandziukCognitivelyPlausibleGame2011, mcilroy-youngAligningSuperhumanAI2020}, while retaining search as a component of decision making.

Further experiments should extend the evaluation to a wider range of games (e.g. on Ludii \cite{Piette_2020_Ludii}) and explore additional configurations of playout budgets, node pool sizes, and GRAVE parameters. An open question is whether the combination of two-level search and node recycling yields comparable or greater improvements in other MCTS variants, such as HRAVE \cite{sironiComparisonRapidAction2016} or ISMCTS \cite{cowlingInformationSetMonte2012}.

%\begin{credits}
%\subsubsection{\ackname} A bold run-in heading in small font size at the end of the paper is
%used for general acknowledgments, for example: This study was funded
%by X (grant number Y).

%\subsubsection{\discintname}
%It is now necessary to declare any competing interests or to specifically
%state that the authors have no competing interests. Please place the
%statement with a bold run-in heading in small font size beneath the
%(optional) acknowledgments\footnote{If EquinOCS, our proceedings submission
%system, is used, then the disclaimer can be provided directly in the system.},
%for example: The authors have no competing interests to declare that are
%relevant to the content of this article. Or: Author A has received research
%grants from Company W. Author B has received a speaker honorarium from
%Company X and owns stock in Company Y. Author C is a member of committee Z.
%\end{credits}
%
% ---- Bibliography ----
%
% BibTeX users should specify bibliography style 'splncs04'.
% References will then be sorted and formatted in the correct style.
%
\bibliographystyle{splncs04}
\bibliography{refs}

@inproceedings{cazenaveGeneralizedRapidAction,
  author    = {Tristan Cazenave},
  title     = {Generalized Rapid Action Value Estimation},
  booktitle = {Proceedings of the Twenty-Fourth International Joint Conference on
               Artificial Intelligence, {IJCAI} 2015},
  pages     = {754--760},
  year      = {2015}
}

@phdthesis{breukerMemorySearchGames1998,
title = "Memory versus search in games",
author = "D.M. Breuker",
year = "1998",
language = "English",
isbn = "9090120068",
publisher = "Universiteit Maastricht",
}

@book{allisSearchingSolutionsGames1994,
  title = {Searching for Solutions in Games and Artificial Intelligence},
  author = {Allis, L.V.},
  year = 1994,
}

@inproceedings{powleyMemoryBoundedMonte2017,
  title = {Memory {{Bounded Monte Carlo Tree Search}}},
  booktitle = {Proceedings of the {{Thirteenth AAAI Conference}} on {{Artificial Intelligence}} and {{Interactive Digital Entertainment}}},
  author = {Powley, Edward Jack and Cowling, Peter I. and Whitehouse, Daniel},
  year = 2017,
  pages = {94--100},
  publisher = {AAAI Press},
}

@inproceedings{coulomEfficientSelectivityBackup2006,
  title = {Efficient {{Selectivity}} and {{Backup Operators}} in {{Monte-Carlo Tree Search}}},
  booktitle = {Computers and {{Games}}, 5th {{International Conference}}, {{CG}} 2006},
  author = {Coulom, R{\'e}mi},
  year = 2006,
  series = {Lecture {{Notes}} in {{Computer Science}}},
  volume = {4630},
  pages = {72--83},
  publisher = {Springer},
}

@inproceedings{helmboldAllMovesAsFirstHeuristicsMonteCarlo2009,
  title = {All-{{Moves-As-First Heuristics}} in {{Monte-Carlo Go}}},
  booktitle = {Proceedings of the 2009 {{International Conference}} on {{Artificial Intelligence}}, {{ICAI}} 2009, {{July}} 13-16, 2009, 2 {{Volumes}}},
  author = {Helmbold, David P. and {Parker-Wood}, Aleatha},
  year = 2009,
  pages = {605--610},
  publisher = {CSREA Press}
}

@inproceedings{bouzy2003MonteCarloDevelopments,
  author       = {Bruno Bouzy and
                  Bernard Helmstetter},
  title        = {Monte-Carlo Go Developments},
  booktitle    = {Advances in Computer Games, Many Games, Many Challenges, 10th International
                  Conference, {ACG} 2003, Graz, Austria},
  series       = {{IFIP}},
  volume       = {263},
  pages        = {159--174},
  publisher    = {Kluwer},
  year         = {2003},
}

@inproceedings{kocsisBanditBasedMonteCarlo2006,
  title = {Bandit {{Based Monte-Carlo Planning}}},
  booktitle = {Machine {{Learning}}: {{ECML}} 2006, 17th {{European Conference}} on {{Machine Learning}}, 2006, {{Proceedings}}},
  author = {Kocsis, Levente and Szepesv{\'a}ri, Csaba},
  year = 2006,
  series = {Lecture {{Notes}} in {{Computer Science}}},
  volume = {4212},
  pages = {282--293},
  publisher = {Springer},
}

@article{gellyMonteCarloTreeSearch2011,
  title = {Monte-{{Carlo}} Tree Search and Rapid Action Value Estimation in Computer {{Go}}},
  author = {Gelly, Sylvain and Silver, David},
  year = 2011,
  journal = {Artif. Intell.},
  volume = {175},
  number = {11},
  pages = {1856--1875},
}

@misc{rautureau2025cogniplayworkinprogresshumanlikemodel,
      title={CogniPlay: a work-in-progress Human-like model for General Game Playing}, 
      author={Aloïs Rautureau and Éric Piette},
      year={2025},
      eprint={2507.05868},
      archivePrefix={arXiv},
      primaryClass={cs.AI},
      url={https://arxiv.org/abs/2507.05868}, 
}

@article{agresti1998,
  title={Approximate is better than “exact” for interval estimation of binomial proportions},
  author={Agresti, Alan and Coull, Brent A},
  journal={The American Statistician},
  volume={52},
  number={2},
  pages={119--126},
  year={1998},
  publisher={Taylor \& Francis}
}

@incollection{atkinsonHumanMemoryProposed1968,
  title = {Human {{Memory}}: {{A Proposed System}} and Its {{Control Processes}}},
  shorttitle = {Human {{Memory}}},
  booktitle = {Psychology of {{Learning}} and {{Motivation}}},
  author = {Atkinson, Richard C. and Shiffrin, Richard M.},
  year = 1968,
  series = {Psychology of {{Learning}} and {{Motivation}}},
  volume = {2},
  pages = {89--195},
  publisher = {Elsevier},
}

@article{millerMagicalNumberSeven1956,
  title = {The Magical Number Seven, plus or Minus Two: {{Some}} Limits on Our Capacity for Processing Information},
  shorttitle = {The Magical Number Seven, plus or Minus Two},
  author = {Miller, George A.},
  year = 1956,
  journal = {Psychological Review},
  volume = {63},
  number = {2},
  pages = {81--97},
  publisher = {American Psychological Association},
  address = {US},
  issn = {1939-1471},
}

@inproceedings{Koriche_2017_Symmetry,
	title        = {Constraint-Based Symmetry Detection in General Game Playing},
	author       = {F. Koriche and S. Lagrue and {\'E}. Piette and S. Tabary},
	year         = 2017,
	booktitle    = {Proceedings of the Twenty-Sixth International Joint Conference on Artificial Intelligence, {IJCAI-17}},
	pages        = {280--287}
}

@book{degrootThoughtChoiceChess1946,
  title = {Thought and {{Choice}} in {{Chess}}},
  author = {De Groot, A. D.},
  year = 1946,
  pages = {315},
  publisher = {Noord-Holl. Uitg.},
}

@inproceedings{finnssonLearningSimulationControl2010,
  title = {Learning {{Simulation Control}} in {{General Game-Playing Agents}}},
  booktitle = {Proceedings of the {{Twenty-Fourth AAAI Conference}} on {{Artificial Intelligence}}, {{AAAI}} 2010, {{USA}}, {{July}} 11-15, 2010},
  author = {Finnsson, Hilmar and Bj{\"o}rnsson, Yngvi},
  year = 2010,
  pages = {954--959},
  publisher = {AAAI Press},
}

@article{Koriche_2017_Woodstock,
title = {WoodStock : Un programme-joueur générique dirigé par les contraintes stochastiques},
author = {F. Koriche and S. Lagrue and {\'E}. Piette and S. Tabary},
year = 2017,
journal = "Revue D'Intelligence Artificielle (RIA)",
volume = "31",
pages = "307--336",
number = "3"
}

@inproceedings{winandsPNPN2PN2001,
title = {{{PN}}, {{PN2}} and {{PN*}} in Lines of Action},
author = "M.H.M. Winands and J.W.H.M. Uiterwijk",
year = "2001",
booktitle = "The CMG Sixth Computer Olympiad Computer-Games Workshop Proceedings",
publisher = "Technical Reports in Computer Science CS 01-04, Universiteit Maastricht",
}

@inproceedings{Piette_2020_Ludii,
        author      = "{\'E}. Piette and D. J. N. J. Soemers and M. Stephenson and C. F. Sironi and M. H. M. Winands and C. Browne",
        booktitle   = "ECAI 2020",
        title       = "Ludii -- The Ludemic General Game System",
        pages       = "411-418",
        year        = "2020",
        volume      = "325",
        series      = "Frontiers in Artificial Intelligence and Applications",
}

@inproceedings{mcilroy-youngAligningSuperhumanAI2020,
  title = {Aligning {{Superhuman AI}} with {{Human Behavior}}: {{Chess}} as a {{Model System}}},
  booktitle = {{{KDD}} '20: {{The}} 26th {{ACM SIGKDD Conference}} on {{Knowledge Discovery}} and {{Data Mining}}, {{Virtual Event}}, 2020},
  author = {{McIlroy-Young}, Reid and Sen, Siddhartha and Kleinberg, Jon M. and Anderson, Ashton},
  year = 2020,
  pages = {1677--1687},
  publisher = {ACM},
}

@article{mandziukCognitivelyPlausibleGame2011,
  title = {Towards {{Cognitively Plausible Game Playing Systems}}},
  author = {Mandziuk, Jacek},
  year = 2011,
  journal = {IEEE Comput. Intell. Mag.},
  volume = {6},
  number = {2},
  pages = {38--51},
}

@phdthesis{pietteNouvelleApprocheAu2016,
  title = {Une Nouvelle Approche Au {{General Game Playing}} Dirig\'ee Par Les Contraintes},
  author = {Piette, {\'E}ric},
  year = 2016,
  school = {Artois University, Arras, France}
}

@inproceedings{sironiComparisonRapidAction2016,
  title = {Comparison of Rapid Action Value Estimation Variants for General Game Playing},
  booktitle = {{{IEEE Conference}} on {{Computational Intelligence}} and {{Games}}, {{CIG}} 2016},
  author = {Sironi, Chiara F. and Winands, Mark H. M.},
  year = 2016,
  pages = {1--8},
  publisher = {IEEE},
}

@article{cowlingInformationSetMonte2012,
  title = {Information {{Set Monte Carlo Tree Search}}},
  author = {Cowling, Peter I. and Powley, Edward Jack and Whitehouse, Daniel},
  year = 2012,
  journal = {IEEE Trans. Comput. Intell. AI Games},
  volume = {4},
  number = {2},
  pages = {120--143},
}

@inproceedings{moriartyEvolvingNeuralNetworks1994,
  title = {Evolving {{Neural Networks}} to {{Focus Minimax Search}}},
  booktitle = {Proceedings of the 12th {{National Conference}} on {{Artificial Intelligence}}, {{Volume}} 2},
  author = {Moriarty, David E. and Miikkulainen, Risto},
  year = 1994,
  pages = {1371--1377},
  publisher = {AAAI Press / The MIT Press},
}

\end{document}